\documentclass[11pt,a4paper]{article}
\usepackage[hyperref]{acl2021}
\usepackage{times}
\usepackage{latexsym}

\usepackage{microtype}

\aclfinalcopy 
\def\aclpaperid{934} 

\usepackage{siunitx}
\usepackage{booktabs}
\usepackage{enumitem}
\usepackage{graphicx}
\usepackage{amsmath}
\usepackage{amsfonts}
\usepackage{caption}
\usepackage{subcaption}

\title{Robust Knowledge Graph Completion with Stacked Convolutions and a Student Re-Ranking Network}

\author{Justin Lovelace$^1$ \quad Denis Newman-Griffis$^2$ \quad Shikhar Vashishth$^{3}$\thanks{\, \ Work performed while at Carnegie Mellon University.}\\ \textbf{Jill Fain Lehman}$^4$ \quad \textbf{Carolyn Penstein Ros{\'e}}$^1$ \\ 
$^{1}$Language Technologies Institute, Carnegie Mellon University, USA\\
$^{2}$Department of Biomedical Informatics, University of Pittsburgh, USA\\
$^{3}$Microsoft Research, \\
$^{4}$Human-Computer Interaction Institute, Carnegie Mellon University, USA\\
\small{\texttt{\{jlovelac, jfl, cpa3\}@cs.cmu.edu}, \texttt{dnewmangriffis@pitt.edu}} \\ \small{\texttt{t-svashishth@microsoft.com}}
}

\begin{document}
\newcommand{\refalg}[1]{Algorithm \ref{#1}}
\newcommand{\refeqn}[1]{Equation \ref{#1}}
\newcommand{\reffig}[1]{Figure \ref{#1}}
\newcommand{\reftbl}[1]{Table \ref{#1}}
\newcommand{\refsec}[1]{Section \ref{#1}}

\newcommand{\datasmd}{SNOMED CT Core}
\newcommand{\datafbs}{FB15k-237-Sparse}
\newcommand{\datacon}{CN-100K}
\newcommand{\datafb}{FB15k-237}
\newcommand{\datawn}{WN18RR}
\newcommand{\datayago}{YAGO3-10}

\newcommand{\reminder}[1]{\textcolor{red}{[[ #1 ]]}\typeout{#1}}
\newcommand{\reminderR}[1]{\textcolor{gray}{[[ #1 ]]}\typeout{#1}}

\newcommand{\add}[1]{\textcolor{red}{#1}\typeout{#1}}
\newcommand{\remove}[1]{\sout{#1}\typeout{#1}}

\newcommand{\tensor}{\mathcal{X}}
\newcommand{\real}{\mathbb{R}}

\newcommand{\tuples}{\mathbb{T}}

\newcommand{\argmax}{arg\,max}

\newcommand\norm[1]{\left\lVert#1\right\rVert}

\newcommand{\note}[1]{\textcolor{blue}{#1}}

\newcommand*{\Scale}[2][4]{\scalebox{#1}{$#2$}}%
\newcommand*{\Resize}[2]{\resizebox{#1}{!}{$#2$}}%
\definecolor{officegreen}{rgb}{0.0, 0.5, 0.0}
\def\mat#1{\mbox{\bf #1}}
\maketitle
\begin{abstract}
Knowledge Graph (KG) completion research usually focuses on densely connected benchmark datasets that are not representative of real KGs. We curate two KG datasets that include biomedical and encyclopedic knowledge and use an existing commonsense KG dataset to explore KG completion in the more realistic setting where dense connectivity is not guaranteed. We develop a deep convolutional network that utilizes textual entity representations and demonstrate that our model outperforms recent KG completion methods in this challenging setting. We find that our model's performance improvements stem primarily from its robustness to sparsity. We then distill the knowledge from the convolutional network into a student network that re-ranks promising candidate entities. This re-ranking stage leads to further improvements in performance and demonstrates the effectiveness of entity re-ranking for KG completion.\footnote{\url{https://github.com/justinlovelace/robust-kg-completion}}
\end{abstract}

\section{Introduction}

Knowledge graphs (KGs) have been shown to be useful for a wide range of NLP tasks, such as question answering \cite{kg_in_qa1,kg_in_qa2}, dialog systems \citep{kg_in_dialog}, relation extraction \citep{ds_begin,reside}, and recommender systems \citep{kb_recommender}. However, because scaling the collection of facts to provide coverage for all the true relations that hold between entities is difficult, most existing KGs are incomplete \cite{kg_incomplete}, limiting their utility for downstream applications. 
Because of this problem, KG completion (KGC) has come to be a widely studied task \citep{distmult, complex, convtranse, dettmers2018conve, sun2018rotate, balazevic-etal-2019-tucker, malaviya2020commonsense, interacte2020}. 

The increased interest in KGC has led to the curation of a number of benchmark datasets such as FB15K \citep{bordes2013translating}, WN18 \citep{bordes2013translating}, \datafb{} \citep{FB15K-237}, and YAGO3-10 \citep{rebele2016yago} that have been the focus of most of the work in this area. However, these benchmark datasets are often curated in such a way as to produce densely connected networks that simplify the task and are not representative of real KGs. For instance, FB15K includes only entities with at least 100 links in Freebase, while YAGO3-10 is limited to only include entities in YAGO3 \citep{rebele2016yago} that have at least 10 relations.

Real KGs are not as uniformly dense as these benchmark datasets and have many sparsely connected entities \cite{pujara_sparsity}. This can pose a challenge to typical KGC methods that learn entity representations solely from the knowledge that already exists in the graph. 

Textual entity identifiers can be used to develop entity embeddings that are more robust to sparsity \citep{malaviya2020commonsense}. It has also been shown that textual triplet representations can be used with BERT for triplet classification \citep{kg-bert}. Such an approach can be extended to the more common ranking paradigm through the exhaustive evaluation of candidate triples, but that does not scale to large KG datasets. 

In our work, we found that existing neural KGC models lack the complexity to effectively fit the training data when used with the pre-trained textual embeddings that are necessary for representing sparsely connected entities. We develop an expressive deep convolutional model that utilizes textual entity representations more effectively and improves sparse KGC. We also develop a student re-ranking model that is trained using knowledge distilled from our original ranking model and demonstrate that the re-ranking procedure is particularly effective for sparsely connected entities. Through these innovations, we develop a KGC pipeline that is more robust to the realities of real KGs. Our contributions can be summarized as follows.

\begin{itemize}[itemsep=2pt,parsep=0pt,partopsep=0pt,leftmargin=*,topsep=0.2pt]
    \item We develop a deep convolutional architecture that utilizes textual embeddings more effectively than existing neural KGC models and significantly improves performance for sparse KGC.
    \item We develop a re-ranking procedure that distills knowledge from our ranking model into a student network that re-ranks promising candidate entities. 
    \item We curate two sparse KG datasets containing biomedical and encyclopedic knowledge to study KGC in the setting where dense connectivity is not guaranteed. We release the encyclopedic dataset and the code to derive the biomedical dataset to encourage future work.
\end{itemize}


\section{Related Work}
\textbf{Knowledge Graph Completion: }
KGC models typically learn entity and relation embeddings based on known facts \citep{rascal,bordes2013translating,distmult} and use the learned embeddings to score potential candidate triples. Recent work includes both non-neural \cite{hole,complex,analogy,sun2018rotate} and neural \cite{neural_tensor_network,kg_incomplete,dettmers2018conve,compgcn} approaches for embedding KGs. However, most of them only demonstrate their efficacy on artificially dense benchmark datasets. \citet{pujara_sparsity} show that the performance of such methods varies drastically with sparse, unreliable data. We compare our proposed method against the existing approaches in a realistic setting where the KG is not uniformly dense. 



Prior work has effectively utilized entity names or descriptions to aid KGC \citep{neural_tensor_network, xie2016, Xiao2016SSP}. In more recent work,  \citet{malaviya2020commonsense} explore the problem of KGC using commonsense KGs, which are much sparser than standard benchmark datasets. They adapt an existing KGC model to utilize BERT \citep{devlin-etal-2019-bert} embeddings. In this paper, we develop a deep convoluational architecture that is more effective than adapting existing shallow models which we find to be underpowerered for large KG datasets. 

\citet{kg-bert} developed a triplet classification model by directly fine-tuning BERT with textual entity representations and reported strong classification results. They also adapted their triplet classification model to the ranking paradigm by exhaustively evaluating all possible triples for a given query, $(e_1, r, ?)$. However, the ranking performance was not competitive\footnote{Their reported Hits@10 for FB15K-237 was $.420$ which is lower than all of the models evaluated in this work.}, and such an approach is not scalable to large KG datasets like those explored in this work. Exhaustively applying BERT to compute all rankings for the test set for our largest dataset would take over two months. In our re-ranking setting, we reduce the number of triples that need to be evaluated by over $7700 \times$, reducing the evaluation time to less than 15 minutes.

\textbf{BERT as a Knowledge Base: } Recent work \citep{petroni-etal-2019-language, jiang-etal-2020-know, bertology} has utilized the masked-language-modeling (MLM) objective to probe the knowledge contained within pre-trained models using fill-in-the-blank prompts (e.g. ``Dante was born in \texttt{[MASK]}"). This body of work has found that pre-trained language models such as BERT capture some of the relational knowledge contained within their pre-training corpora. This motivates us to utilize these models to develop entity representations that are well-suited for KGC.

\textbf{Re-Ranking:} \citet{cascade_reranking} introduced cascade re-ranking for document retrieval. This approach applies inexpensive models to develop an initial ranking and utilizes expensive models to improve the ranking of the top-k candidates. Re-ranking has since been successfully applied across many retrieval tasks \citep{qa_reranking, personal-rerank, nogueira2019passage}. Despite re-ranking's widespread success, recent KGC work utilizes a single ranking model. We develop an entity re-ranking procedure and demonstrate the effectiveness of the re-ranking paradigm for KGC.

\textbf{Knowledge Distillation:} Knowledge distillation is a popular technique that is often used for model compression where a large, high-capacity teacher is used to train a simpler student network \citep{hinton-kd}. However, knowledge distillation has since been shown to be useful for improving model performance beyond the original setting of model compression.  \citet{noisy-kd} demonstrated that knowledge distillation improved image classification performance in a setting with noisy labels. The incompleteness of KGs leads to noisy training labels which motivates us to use knowledge distillation to train a student re-ranking model that is more robust to the label noise.

\begin{table}[t]
\centering
 \resizebox{\linewidth}{!}{%
	\begin{tabular}{lrcrrr}
	\toprule
	Dataset        & \multicolumn{1}{c}{\textbf{\# Nodes}}   & \multicolumn{1}{c}{\textbf{\# Rels}}     & \multicolumn{1}{c}{\textbf{\# Train}} & \multicolumn{1}{c}{\textbf{\# Valid}} & \multicolumn{1}{c}{\textbf{\# Test}} \\
	\midrule
	\datafb{} & 14,451   & 237   & 272,115    & 17,535 & 20,466        \\
	\datasmd{} & 77,316   & 140  & 502,224     & 71,778  & 143,486       \\ 
	\datacon{} & 78,088   & 34   & 100,000    & 1,200  & 1,200       \\ 
	\datafbs{} & 14,451  & 237   & 18,506     & 17,535 & 20,466         \\ 
	\bottomrule
	\end{tabular}
 }%
\caption{\label{tab:dataset_stats}Dataset statistics}

\end{table}

\begin{figure}[t]
 \centering
 \includegraphics[width=\linewidth]{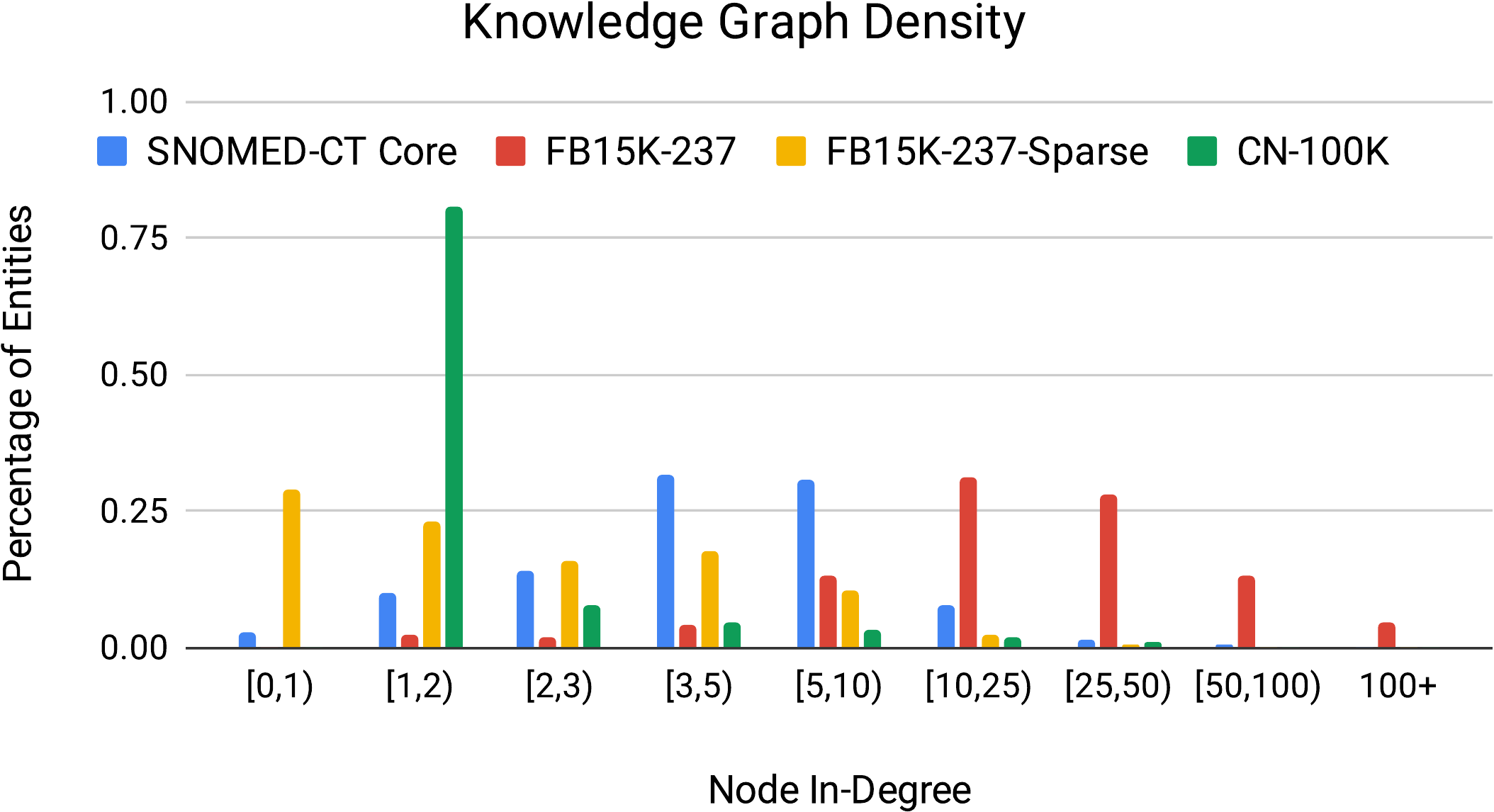}
 \caption{In-degrees of entities in the training KGs (including inverse relations) \label{fig:sparsity_training_kgs}}
\end{figure}

\section{Datasets}
We examine KGC in the realistic setting where KGs have many sparsely connected entities. 
We utilize a commonsense KG dataset that has been used in past work and curate two additional sparse KG datasets containing biomedical and encyclopedic knowledge. We release the encyclopedic dataset and the code to derive the biomedical dataset to encourage future work in this challenging setting.
The summary statistics for all datasets are presented in Table~\ref{tab:dataset_stats} and we visualize the connectivity of the datasets in Figure~\ref{fig:sparsity_training_kgs}. 

\subsection{\datasmd{}}
For constructing \datasmd{}, we use the knowledge graph defined by SNOMED CT \citep{snomed}, which is contained within the Unified Medical Language System (UMLS) \citep{Bodenreider2004}.
SNOMED CT is well-maintained and is one of the most comprehensive knowledge bases contained within the UMLS \cite{Jimenez-Ruiz2011, 10.1197/jamia.M2541}. We first extract the UMLS\footnote{We work with the 2020AA release of the UMLS.} concepts found in the CORE Problem List Subset of the SNOMED CT knowledge base. This subset is intended to contain the concepts most useful for documenting clinical information. We then expand the graph to include all concepts that are directly linked to those in the CORE Problem List Subset according to the relations defined by the SNOMED CT KG. Our final KG consists of this set of concepts and the SNOMED CT relations connecting them. Importantly, we do not filter out rare entities from the KG, as is commonly done during the curation of benchmark datasets.

To avoid leaking data from inverse, or otherwise informative, relations, we divide the facts into training, validation, and testing sets based on unordered tuples of entities $\{e_1, e_2\}$ so that all relations between any two entities are confined to a single split. Unlike some other KG datasets that filter out inverse relations, we divide our dataset in such a way that this is not necessary; our dataset already includes inverse relations, and they do not need to be manually added for training and evaluation as is standard practice \citep{dettmers2018conve, malaviya2020commonsense}.

Because we represent entities using textual descriptions in this work, we also mine the entities' preferred concept names (e.g. \textit{``Traumatic hematoma of left kidney''}) from the UMLS.

\begin{figure*}[t]
    \centering
    \includegraphics[width=\textwidth]{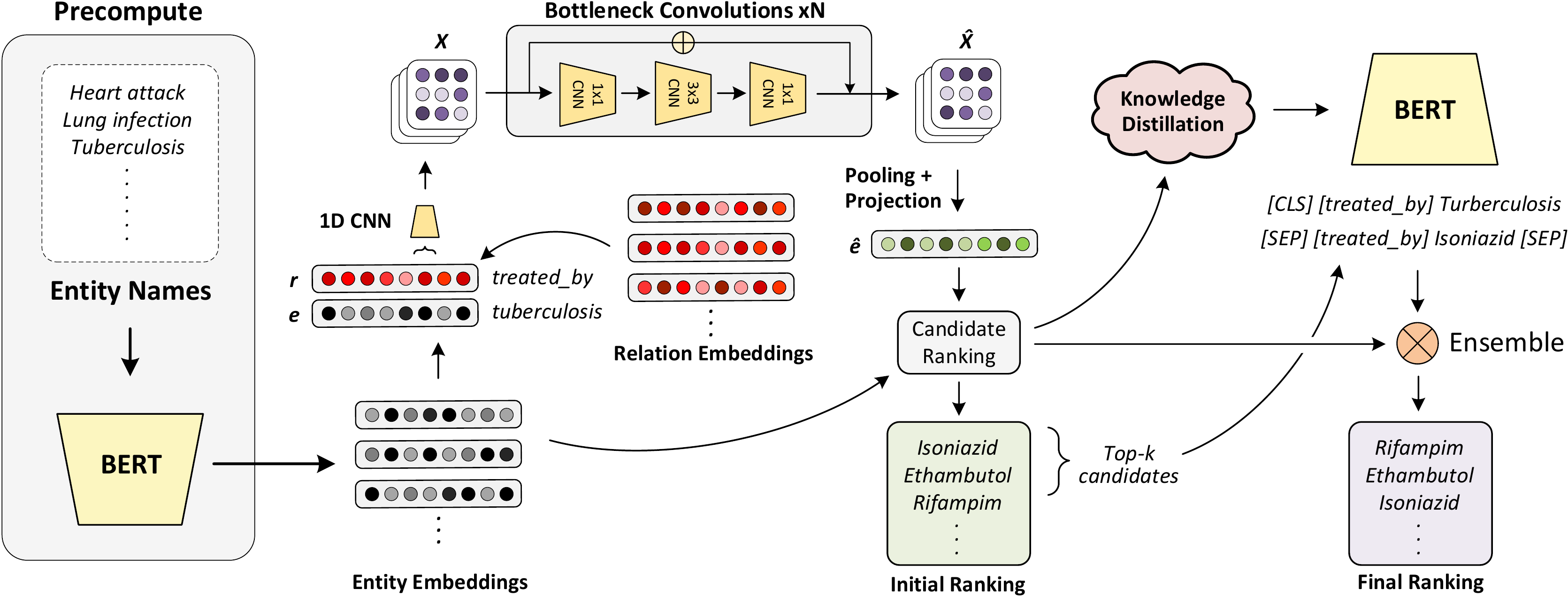}
    \caption{We utilize BERT to precompute entity embeddings. We then stack the precomputed entity embedding with a learned relation embedding and project them to a two-dimensional spatial feature map, upon which we apply a sequence of two-dimensional convolutions. The final feature map is then average pooled and projected to a query vector, which is used to rank candidate entities. We extract promising candidates and train a re-ranking model utilizing knowledge distilled from the original ranking model. The final candidate ranking is generated by ensembling the ranking and re-ranking models.}
    \label{fig:model_arch}
\end{figure*}

\subsection{\datafbs{}}
The \datafb{} \citep{FB15K-237} dataset contains encyclopedic knowledge about the world, e.g. \textit{(Barack Obama, placeOfBirth, Honolulu)}.
Although the dataset is very densely connected, that density is artificial. FB15K \citep{bordes2013translating}, the precursor to \datafb{}, was curated to only include entities with at least 100 links in Freebase \cite{freebase}. 

The dense connectivity of \datafb{} does allow us to to ablate the effect of this density. We utilize the \datafb{} dataset and also develop a new dataset, denoted \datafbs{}, by randomly downsampling the facts in the training set of \datafb{} to match the average in-degree of the ConceptNet-100K dataset. We use this to directly evaluate the effect of increased sparsity.

For the \datafb{} dataset, we use the textual identifiers released by \citet{xie2016}. They released both entity names (e.g. ``Jason Frederick Kidd'') as well as brief textual descriptions (e.g. ``Jason Frederick Kidd is a retired American professional basketball player\dots'') for most entities. We utilize the textual descriptions when available.

\subsection{ConceptNet-100K}
ConceptNet \citep{ConceptNet} is a KG that contains commonsense knowledge about the world such as the fact \textit{(go to dentist, motivatedBy, prevent tooth decay)}. We utilize ConceptNet-100k (\datacon{}) \citep{li-etal-2016-commonsense} which consists of the Open Mind Common Sense entries in the ConceptNet dataset. This KG is much sparser than benchmark datasets like \datafb{}, which makes it well-suited for our purpose. We use the training, validation, and testing splits of \citet{malaviya2020commonsense} to allow for direct comparison. We also use the textual descriptions released by \citet{malaviya2020commonsense} to represent the KG entities.

\section{Methods}
We provide an overview of our model architecture in Figure~\ref{fig:model_arch}. We first extract feature representations from BERT \citep{devlin-etal-2019-bert} to develop textual entity embeddings. Motivated by our observation that existing neural KG architectures are underpowered in our setting, we develop a deep convolutional network utilizing architectural innovations from deep convolutional vision models. Our model's design improves its ability to fit complex relationships in the training data which leads to downstream performance improvements. 

Finally, we distill our ranking model's knowledge into a student re-ranking network that adjusts the rankings of promising candidates. In doing so, we demonstrate the effectiveness of the re-ranking paradigm for KGC and develop a KGC pipeline with greater robustness to the sparsity of real KGs.

\subsection{Entity Ranking}
\label{sec:entity_ranking}
We follow the standard formulation for KGC. We represent a KG as a set of entity-relation-entity facts $(e_1,r,e_2)$. Given an incomplete fact, $(e_1,r,?)$, our model computes a score for all candidate entities $e_i$ that exist in the graph. An effective KGC model should assign greater scores to correct entities than incorrect ones. We follow recent work \citep{dettmers2018conve, malaviya2020commonsense} and consider both forward and inverse relations (e.g. treats and treated\_by) in this work. For the datasets that do not already include inverse relations, we introduce an inverse fact, $(e_2,r^{-1},e_1)$, for every fact, $(e_1,r,e_2)$, in the dataset.

\subsubsection{Textual Entity Representations}
\label{sec:ent_rep}
We utilize BERT \citep{devlin-etal-2019-bert} to develop entity embeddings that are invariant to the connectivity of the KG. 
We follow the work of \citet{malaviya2020commonsense} and adapt BERT to each KG's naming style by fine-tuning BERT using the MLM objective with the set of entity identifiers in the KG.

For \datacon{} and \datafb{}, we utilize the BERT-base uncased model. For \datasmd{} KG, we utilize PubMedBERT \citep{pubmedbert} which is better suited for the biomedical terminology in the UMLS.

We apply BERT to the textual entity identifiers and mean-pool across the token representations from all BERT layers to obtain a summary feature vector for the concept name. We fix these embeddings during training because we must compute scores for a large number of potential candidate entities for each training example. This makes fine-tuning BERT prohibitively expensive.

\subsubsection{Deep Convolutional Architecture}
Inspired by the success of deep convolutional models in computer vision \citep{deepconv, vgg, resnet, huang2019convolutional, huang2017densely}, we develop a knowledge base completion model based on the seminal ResNet architecture \citep{resnet} that is sufficiently expressive to model complex interactions between the BERT feature space and the relation embeddings.

Given an incomplete triple $(e_i, r_j, ?)$, we begin by stacking the precomputed entity embedding $\mathbf{e} \in \mathbb{R}^{1 \times d}$ with the learned relation embedding of the same dimension $\mathbf{r} \in \mathbb{R}^{1 \times d}$ to produce a feature vector of length $d$ with two channels $\mathbf{q} \in \mathbb{R}^{2 \times d}$.
We then apply a one-dimensional convolution with a kernel of width $1$ along the length of the feature vector to project each position $i$ to a two-dimensional spatial feature map $\mathbf{x_i} \in \mathbb{R}^{f \times f}$ where the convolution has $f \times f$ filters. Thus the convolution produces a two-dimensional spatial feature map $\mathbf{X} \in \mathbb{R}^{f \times f \times d}$ with $d$ channels, representing the incomplete query triple $(e_i, r_j, ?)$.

The spatial feature map, $\mathbf{X} \in \mathbb{R}^{f \times f \times d}$, is analogous to a square image with a side length of $f$ and $d$ channels, allowing for the straightforward application of deep convolutional models such as ResNet. We apply a sequence of $3N$ bottleneck blocks to the spatial feature map where $N$ is a hyperparameter that controls the depth of the network. A bottleneck block consists of three consecutive convolutions: a $1\times1$ convolution, a $3\times3$ convolution, and then another $1\times1$ convolution. The first $1 \times 1$ convolution reduces the feature map dimensionality by a factor of $4$ and then the second $1 \times 1$ convolution restores the feature map dimensionality. This design reduces the dimensionality of the expensive $3 \times 3$ convolutions and allows us to increase the depth of our model without dramatically increasing its parameterization. We double the feature dimensionality of the bottleneck blocks after $N$ and $2N$  blocks so the dimensionality of the final feature map produced by the sequence of convolutions is $4d$.

We add residual connections to each bottleneck block which improves training for deep networks \citep{resnet}. If we let $\mathcal{F}(X)$ represent the application of the bottleneck convolutions, then the output of the bottleneck block is $Y = \mathcal{F}(X) + X$. We apply batch normalization followed by a ReLU nonlinearity \cite{relu} before each convolutional layer \citep{preactivation} .

We utilize circular padding \cite{omnidirectionalwang2018,interacte2020} with the $3\times3$ convolutions to maintain the spatial size of the feature map and use a stride of $1$ for all convolutions. For the bottleneck blocks that double the dimensionality of the feature map, we utilize a projection shortcut for the residual connection \citep{resnet}. 

\subsubsection{Entity Scoring}
Given an incomplete fact $(e_i, r_j, ?)$, our convolutional architecture produces a feature map $\mathbf{\hat{X}} \in \mathbb{R}^{f \times f \times 4d}$. We average pool this feature representation over the spatial dimension which produces a summary feature vector $\mathbf{\hat{x}} \in \mathbb{R}^{4d}$. We then apply a fully connected layer followed by a PReLU nonlinearity \citep{prelu} to project the feature vector back to the original embedding dimensionality $d$. We denote this final vector 
$\mathbf{\hat{e}}$ 
and compute scores for candidate entities using the dot product with candidate entity embeddings. The scores can be efficiently computed for all entities simultaneously using a matrix-vector product with the embedding matrix $\mathbf{y} = \mathbf{\hat{e}} \mathbf{E}^T$ where $\mathbf{E}\in \mathbb{R}^{m \times d}$ stores the embeddings for all $m$ entities in the KG.

\subsubsection{Training}
Adopting the terminology used by \citet{Ruffinelli2020You}, we utilize a 1vsAll training strategy with the binary cross-entropy loss function. We treat every fact in our dataset, $(e_i, r_j, e_k)$, as a training sample where $(e_i, r_j, ?)$ is the input to the model. We compute scores for all entities as described previously and apply a sigmoid operator to induce a probability for each entity. We treat all entities other than $e_k$ as negative candidates and then compute the binary cross-entropy loss.

We train our model using the Adam optimizer \citep{adam} with decoupled weight decay regularization \citep{adamw} and label smoothing. We train our models for a maximum of 200 epochs and terminate training early if the validation Mean Reciprocal Rank (MRR) has not improved for 20 epochs. We trained all of the models used in this work using a single NVIDIA GeForce GTX 1080 Ti.

\subsection{Entity Re-Ranking}

\subsubsection{Re-Ranking Network}
We use our convolutional network to extract the top-$k$ entities for every unique training query and then train a re-ranking network to rank these entities. 
We design our student re-ranking network as a triplet classification model that utilizes the full candidate fact, $(e_i, r_j, e_k)$, instead of an incomplete fact, $(e_i, r_j, ?)$. This allows the network to model interactions between all elements of the triple. 
The re-ranking setting also enables us to directly fine-tune BERT which often improves performance \citep{peters-etal-2019-tune}. 

We introduce relation tokens\footnote{We use relation tokens instead of free-text relation representations because the relation identifiers for our datasets are not all well-formed using natural language, and the different styles would introduce a confounding factor that would complicate our evaluation. Utilizing appropriate free-text relation identifiers may improve performance, but we leave that to future work. } for each relation in the knowledge graph and construct the textual input by prepending the head and tail entities with the relation token and then concatenating the two sequences. Thus the triple (``head name'', $r_i$, ``tail name'') would be represented as ``\texttt{[CLS] [REL\_i] head name [SEP] [REL\_i] tail name [SEP]}''. We use a learned linear combination of the \texttt{[CLS]} embedding from each layer as the final feature representation for the prediction.

\subsubsection{Knowledge Distillation}

A sufficiently performant ranking model can provide an informative prior that can be used to smooth the noisy training labels and improve our re-ranking model. For each training query $i$, we normalize the logits produced by our teacher ranking model, $f_T(\mathbf{x}_i)$, for the $k$ candidate triples, $f_T(\mathbf{x}_i)_{0:k}$, as \[\mathbf{s}_{ik:(i+1)k} = \textrm{softmax}(f_T(\mathbf{x}_{i})_{0:k}/T)\] where $T$ is the temperature \citep{hinton-kd}.

Our training objective for our student model, $f_S(x_i)$, is a weighted average of the binary cross entropy loss, $\mathcal{L}_{bce}$, using the teacher's normalized logits, $\mathbf{s}$, and the noisy training labels, $\mathbf{y}$.
\[
\begin{aligned}
\mathcal{L}_{KD}(y_i,x_i) & = \lambda\mathcal{L}_{bce}(s_i,f_S(x_i)) \\
&+ (\lambda -1)\mathcal{L}_{bce}(y_i,f_S(x_i))\\
     & = \mathcal{L}_{bce}((\lambda-1)y_i + \lambda s_i,f_S(x_i)) \\
\end{aligned}
\]
 We select $\lambda \in \{.25, .5, .75, 1\}$ to optimize the balance between the two objectives using validation performance. 

\subsubsection{Training}
For our experiments, we extract the top $k=10$ candidates produced by our ranking model for every query in the training set. We train our student network using the Adam optimizer \citep{adam} with decoupled weight decay regularization \citep{adamw}. We fine-tune BERT for a maximum of 10 epochs and terminate training early if the Mean Reciprocal Rank (MRR) on validation data has not improved for 3 epochs. 

\subsubsection{Student-Teacher Ensemble}
For every query, we apply our re-ranking network to the top $k=10$ triples and compute the final ranking using an ensemble of the teacher and student networks. The final ranking are computed with
\[
\begin{aligned}
\mathbf{\hat{s}}_{ik:(i+1)k} & = \alpha (\textrm{softmax}(f_S(\mathbf{x}_{ik:(i+1)k}))) \\
&+(1-\alpha)(\textrm{softmax}(f_T(\mathbf{x}_i)_{0:k})))\\
\end{aligned}
\]
where $0 \leq \alpha \leq 1$ controls the impact of the student re-ranker. The cost of computing $\mathbf{\hat{s}}_{ik:(i+1)k}$ is negligible, so we sweep over $[0,1]$ in increments of $.01$ and select the $\alpha$ that achieves the best validation MRR.

\begin{table*}[t]
\small
\centering
\resizebox{\textwidth}{!}{
\begin{tabular}{lcccccccccc}
\toprule
       & \multicolumn{5}{c}{\textbf{\datasmd{}}} & \multicolumn{5}{c}{\textbf{\datacon{}}} \\
\cmidrule(r){2-6} \cmidrule(r){7-11}
       & MR & MRR   & H@1     & H@3 & H@10 & MR & MRR   & H@1     & H@3 & H@10 \\
\midrule
DistMult $[\clubsuit]$  & $5146$ & $.293$  & $.226$  & $.318$ & $.426$  & $-$ & $.090$  &  $.045$     & $.098$ & $.174$  \\
ComplEx $[\clubsuit]$ & $3903$ & $.302$  & $.224$  & $.332$ & $.456$ & $-$ & $.114$   & $.074$     & $.125$ & $.190$  \\
ConvE $[\clubsuit]$& $3739$ & $.271$  & $.191$  & $.303$ & $.429$ & $-$  & $.209$   & $.140$     & $.229$ & $.340$ \\
ConvTransE $[\clubsuit]$& $3585$ & $.290$  & $.213$  & $.321$ & $.442$ & $-$ & $.187$   & $.079$     & $.239$ & $.390$  \\
\midrule
BERT-ConvE & $414$ & $.383$  & $.277$  & $.430$ & $.591$ & $260$ & $.453$  & $.332$  & $.521$ & $.691$   \\
BERT-ConvTransE & $514$ & $.373$  & $.273$  & $.417$ & $.568$ & $276$ & $.458$  & $.340$  & $.520$ & $.675$   \\
BERT-Large-ConvTransE $[\clubsuit]$ & $-$ & $-$  & $-$  & $-$ & $-$ & $-$ & $.523$ & $.410$   & $.585$      & $.735$   \\ 
\midrule
BERT-DeepConv & $265$ & $\underline{.479}$  & $.374$  & $.532$ & $.685$ & $\mathbf{161}$ & $\underline{.540}$  & $.418$  & $.610$ & $\mathbf{.772}$   \\
\midrule
BERT-ResNet & $265$ & $\underline{.492}^{\ast}$  & $.389$   & $.544$ & $\mathbf{.691}$& $169$ & $\underline{.550^{\ast}}$  & $.426$  & $.628$ & $.769$  \\ 
\hspace{3mm} + Re-ranking  & $265$ & $.562^{ \dagger}$ & $.482$  & $.608$  & $\mathbf{.691}$  & $170$  & $.377$ & $.216$  & $.437$  & $.769$   \\ 

\hspace{6mm} + Knowledge Distillation (KD)   & $265$ & $.566^{ \dagger}$  & $.487$  & $.614$ & $\mathbf{.691}$  & $169$  & $.528$ & $.402$  & $.603$  & $.769$   \\ 

\hspace{6mm} + Ranking Ensemble (RE)  & $\mathbf{264}$ & ${.576^{\dagger}}$  & $\mathbf{.503}$  & $.619$ & $\mathbf{.691}$  & $169$  & $.555$ & $.438$  & $.623$  & $.769$ \\ 

\hspace{6mm} + KD and RE & $\mathbf{264}$ & $\mathbf{{.577}^{\dagger}}$  & $.501$  & $\mathbf{.623}$ & $\mathbf{.691}$ & $169$ & $\mathbf{{.569}^{\dagger}}$ & $\mathbf{.452}$   & $\mathbf{.647}$  & $.769$ \\ 
\bottomrule
\end{tabular}
}

\vspace*{0.2 cm}

\resizebox{\textwidth}{!}{
\begin{tabular}{lcccccccccc}
\toprule
       & \multicolumn{5}{c}{\textbf{\datafb{}}} & \multicolumn{5}{c}{\textbf{\datafbs{}}} \\
\cmidrule(r){2-6} \cmidrule(r){7-11}
       & MR & MRR   & H@1     & H@3 & H@10 & MR & MRR   & H@1     & H@3 & H@10 \\
\midrule
DistMult $[\spadesuit]$ & $-$   & $.343$  & $-$  & $-$ & $.531$ & $3061$ & $.136$  & $.092$  & $.146$ & $.223$ \\ 
ComplEx $[\spadesuit]$  & $-$  & $.348$  & $-$  & $-$ & $\mathbf{.536}$ & $3333$& $.132$  & $.091$  & $.143$ & $.216$ \\
ConvE $[\spadesuit]$& $-$& $.339$   & $-$ & $-$ & $.521$ & $2263$& $.156$  & $.106$  & $.165$ & $.258$ \\
ConvTransE $[\diamondsuit]$& $-$& $.33$  & $.24$  & $.37$ & $.51$& $2285$ & $.153$  & $.103$  & $.161$ & $.255$ \\
\midrule
BERT-ConvE & $193$ & $.305$  & $.224$  & $.330$ & $.465$ & $408$ & $.190$  & $.128$  & $.200$ & $.315$   \\
BERT-ConvTransE & $211$ & $.296$  & $.218$  & $.321$ & $.449$ & $\mathbf{390}$ & $.188$  & $.127$  & $.199$ & $.310$   \\
\midrule
BERT-DeepConv & $190$ & $\underline{.327}$  & $.246$  & $.354$ & $.488$ & $422$ & $.188$  & $.127$  & $.197$ & $.314$   \\
\midrule
BERT-ResNet  & $186$ & $\underline{.346}^{\ast}$  & $.262$  & $.379$ & $.514$ & $413$ & $.191^{\ast}$ & $.128$   & $.201$  & $\mathbf{.317}$ \\ 
\hspace{3mm} + Re-ranking  & $187$ & $.304$ & $.212$  & $.329$  & $.514$  & $413$ & $.190$ & $.128$ & $.200$  & $\mathbf{.317}$    \\ 

\hspace{6mm} + Knowledge Distillation (KD)   & $187$ & $.310$  & $.220$  & $.334$ & $.514$  & $413$  & ${.197}^{ \dagger}$ & $.135$  & $.209$  & $\mathbf{.317}$  \\ 

\hspace{6mm} + Ranking Ensemble (RE)  & $\mathbf{186}$ & $\mathbf{.354^{ \dagger}}$  & $\mathbf{.270}$  & $\mathbf{.387}$ & $.514$  & $413$  & $\mathbf{.199^{ \dagger}}$  & $\mathbf{.137}$ & $.210$   & $\mathbf{.317}$  \\ 

\hspace{6mm} + KD and RE  & $\mathbf{186}$ & $.353^{\dagger}$  & $.269$  & $.386$ & $.514$ & $413$ & $.198^{\dagger}$ & $.136$   & $\mathbf{.211}$  & $\mathbf{.317}$ \\

\bottomrule
\end{tabular}
}
\caption{\label{tab:results2} Comparison of KGC results across all datasets.  We indicate statistical significance for: (1) Improvements of deep convolutional BERT models over both shallow convolutional BERT models with an underline $(p<0.005)$; (2) Improvements of BERT-ResNet over BERT-DeepConv with a $\ast$ $(p<0.05)$; (3) Improvements of the re-ranking configurations over the original rankings with a $\dagger$ $(p<0.005)$. $[\clubsuit]$ indicates that \datacon{} results are from \citet{malaviya2020commonsense}. $[\spadesuit]$ indicates that \datafb{} results are from \citet{Ruffinelli2020You}. $[\diamondsuit]$ indicates that \datafb{} results are from \citet{convtranse}. Dashes indicate that the metric was not reported by the prior work. \label{tab:results} \small}

\end{table*}

\section{Experiments}
\subsection{Baselines}

We utilize the same representative selection of KG models from \citet{malaviya2020commonsense} as baselines: DistMult \citep{distmult}, ComplEx \citep{complex}  ConvE \citep{dettmers2018conve}, and ConvTransE \citep{convtranse}. This is not an exhaustive selection of all recent KG methods, but a recent replication study by \citet{Ruffinelli2020You} found that the baselines that we use are competitive with the state-of-the-art and often outperform more recent models when trained appropriately. 

We develop additional baselines by adapting the shallow convolutional KGC models to use BERT embeddings to evaluate the benefits of utilizing our proposed convolutional architecture instead of simply repurposing existing KGC models. We refer to these models as BERT-ConvE and BERT-ConvTransE. \citet{malaviya2020commonsense} used BERT embeddings in conjunction with ConvTransE for commonsense KGC, but their model was prohibitively large to reproduce. We refer to their model as BERT-Large-ConvTransE and compare directly against their reported results.

We also develop a deep convolutional baseline, termed BERT-DeepConv, to evaluate the effect of the architectural innovations used in our model. BERT-DeepConv transforms the input embeddings to a spatial feature map like our proposed model, but it then applies a stack of $3 \times 3$ convolutions instead of a sequence of bottleneck blocks with residual connections. We select hyperparameters (detailed in the Appendix) for all of our BERT baselines so that they have a comparable number of trainable parameters to our proposed model. We discuss the size of these models in detail in in Section \ref{sec:size}.

To evaluate the impact of our re-ranking stage, we ablate the use of knowledge distillation and ensembling. Thus we conduct experiments where our re-ranker uses only knowledge distillation, uses only ensembling, and uses neither. 
This means that in the most naive setting, we train the re-ranker using the hard training labels and re-rank the candidates using only the re-ranker.

\subsection{Evaluation}
We report standard ranking metrics: Mean Rank (MR), Mean Reciprocal Rank (MRR), Hits at 1 (H@1), Hits at 3 (H@3), and Hits at 10 (H@10). 
We follow past work and use the filtered setting \cite{bordes2013translating}, removing all positive entities other than the target entity before calculating the target entity's rank.

We utilize paired bootstrap significance testing \citep{berg-kirkpatrick-etal-2012-empirical} with the MRR to validate the statistical significance of improvements. To account for the large number of comparisons being performed, we apply the Holm–Bonferroni method \citep{holm} to correct for multiple hypothesis testing. We define families for the three primary hypotheses that we tested with our experiments. They are as follows: 
(1) The deep convolutional BERT models outperform the shallow convolutional BERT models. 
(2) BERT-ResNet improves upon our BERT-DeepConv baseline.
(3) The re-ranking procedure improves the original rankings.

This selection has the benefit of allowing for a more granular analysis of each conclusion while significantly reducing the number of hypotheses. The first family includes all pairwise comparisons between the two deep convolutional models and the two shallow convolutional models. The second family involves all comparisons between BERT-ResNet and BERT-DeepConv. The third family includes comparisons between all re-ranking configurations and the original rankings. We note that the p-value for each family bounds the strict condition that we report any spurious finding within the family.



\section{Results and Discussion}

\subsection{Ranking Performance}
We report results across all of our datasets in Table \ref{tab:results}. Our ranking model, BERT-ResNet, outperforms the previously published models and our baselines across all of the sparse datasets. We find that for all sparse datasets, the models that use free text entity representations outperform the models that learn the entity embeddings during training. Among the models utilizing textual information, the deep convolutional methods generally outperform the adaptations of existing neural KG models. BERT-ResNet outperforms BERT-DeepConv across all datasets, demonstrating that the architectural innovations do improve downstream performance.

 On the full \datafb{} dataset, our proposed model is able to achieve competitive results compared to strong baselines. However, the focus of this work is not to achieve state-of-the-art performance on densely connected benchmark datasets such as \datafb{}. These results do, however, allow us to observe the outsized impact of sparsity on models that do not utilize textual information.
 
 \subsection{Re-Ranking Performance}
 Re-ranking entities without knowledge distillation or ensembling leads to poor results, degrading the MRR across most datasets. We note that the performance of our re-ranking model could be limited by our use of a pointwise loss function. Further exploration of pairwise or listwise learning learning-to-rank methods is a promising direction for future exploration that could lead to further improvements \citet{ranking}.
 
 The inclusion of either knowledge distillation or ensembling improves performance. Ensembling is particularly important, achieving a statistically significant improvement over the initial rankings across most datasets. Our final setting using both knowledge distillation and ensembling is the only setting to achieve a statistically significant improvement across all four datasets, although using both does not consistently improve performance over ensembling alone. 
 
A plausible explanation for this is that knowledge distillation improves performance by reducing the divergence between the re-ranker and the teacher, but ensembling can already achieve a similar effect by simply increasing the weight of the teacher in the final prediction. We observe that the weight of the teacher is reduced across all four datasets when knowledge distillation is used which would be consistent with this explanation. Knowledge distillation has also been shown to be useful in situations with noisy labels \citep{noisy-kd} which may explain why it was particularly effective for our sparsest dataset, \datacon{}, where training with the hard labels led to particularly poor performance.

\subsection{Effect of Re-Ranking}
\begin{figure}[]
 \centering
 \includegraphics[width=0.5\textwidth]{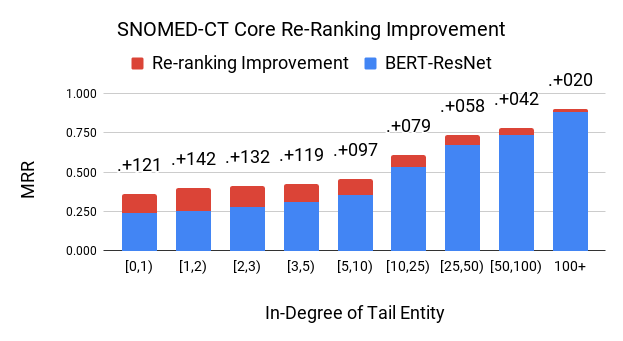}
 \caption{Effect of re-ranking on performance for \datasmd{} across varying levels of sparsity. \label{fig:sparsity_snomed}}
\end{figure}

We bin test examples by the in-degree of the tail nodes and compute the MRR within these bins for our model before and after re-ranking. We report this breakdown for the \datasmd{} dataset in Figure~\ref{fig:sparsity_snomed}. Our re-ranking stage improves performance uniformly across all levels of sparsity, but it is particularly useful for entities that are rarely seen during training. This is also consistent with the comparatively smaller topline improvement for the densely connected \datafb{} dataset. 


\subsection{Model Capacity}
\label{sec:size}
We report the number of trainable parameters for the models that use textual representations along with the train and test set MRR for \datasmd{} in Table~\ref{tab:size}. We observe a monotonic relationship between training and testing performance and note that the shallow models fail to achieve our model's test performance on the training set. This demonstrates that the shallow models lack the complexity to adequately fit the training data. A similar trend held for all datasets except for \datafbs{} whose smaller size reduces the risk of underfitting. This explains the smaller performance improvement for that dataset. 

\begin{table}[]
\centering
\small
\begin{tabular}{lcc}
\toprule
    Model   & Trainable & \multicolumn{1}{c}{\datasmd{} }  \\
       & Params & Train/Test MRR \\
\midrule
BERT-ConvE & $ 34\textrm{M}$  & $.460$ / $.383$ \\ 
BERT-ConvTransE & $ 37\textrm{M}$  & $.449$ /$.373$ \\ 
BERT-DeepConv & $ 38\textrm{M}$  & $.696$ /$.479$     \\ 
BERT-ResNet & $ 33\textrm{M}$  & $.715$ / $.492$ \\ 
\bottomrule
\end{tabular}
\caption{Comparison of trainable parameters for KGC  models  that  utilize  textual  entity representations. \label{tab:size} \small}

\end{table}

\citet{malaviya2020commonsense} scaled up BERT-Large-ConvTransE to use over $524\textrm{M}$ trainable parameters, and their model did outperform our smaller BERT-ConvTransE baseline. However, their model still fails to match the performance of either of our deep convolutional models despite using over $15 \times$ the number of trainable parameters. 




\section{Conclusion}
KGs often include many sparsely connected entities where the use of textual entity embeddings is necessary for strong performance. We develop a deep convolutional network that is better-suited for this setting than existing neural models developed on artificially dense benchmark KGs. We also introduce a re-ranking procedure to distill the knowledge from our convolutional model into a student re-ranking network and demonstrate that our procedure is particularly effective at improving the ranking of sparse candidates. We utilize these innovations to develop a KGC pipeline with greater robustness to the realities of KGs and demonstrate the generalizability of our improvements across biomedical, commonsense, and encyclopedic KGs. 

\section*{Acknowledgments}
This work was supported by the National Science Foundation grant IIS 1917955 and the National Library Medicine of the National Institutes of Health under award number T15 LM007059.

\bibliographystyle{acl_natbib}
\bibliography{anthology,acl2021}

\begin{thebibliography}{63}
\expandafter\ifx\csname natexlab\endcsname\relax\def\natexlab#1{#1}\fi

\bibitem[{Balazevic et~al.(2019)Balazevic, Allen, and
  Hospedales}]{balazevic-etal-2019-tucker}
Ivana Balazevic, Carl Allen, and Timothy Hospedales. 2019.
\newblock \href {https://doi.org/10.18653/v1/D19-1522} {{T}uck{ER}: Tensor
  factorization for knowledge graph completion}.
\newblock In \emph{Proceedings of the 2019 Conference on Empirical Methods in
  Natural Language Processing and the 9th International Joint Conference on
  Natural Language Processing (EMNLP-IJCNLP)}, pages 5185--5194, Hong Kong,
  China. Association for Computational Linguistics.

\bibitem[{Berg-Kirkpatrick et~al.(2012)Berg-Kirkpatrick, Burkett, and
  Klein}]{berg-kirkpatrick-etal-2012-empirical}
Taylor Berg-Kirkpatrick, David Burkett, and Dan Klein. 2012.
\newblock \href {https://www.aclweb.org/anthology/D12-1091} {An empirical
  investigation of statistical significance in {NLP}}.
\newblock In \emph{Proceedings of the 2012 Joint Conference on Empirical
  Methods in Natural Language Processing and Computational Natural Language
  Learning}, pages 995--1005, Jeju Island, Korea. Association for Computational
  Linguistics.

\bibitem[{Bodenreider(2004)}]{Bodenreider2004}
Olivier Bodenreider. 2004.
\newblock The unified medical language system ({UMLS}): integrating biomedical
  terminology.
\newblock \emph{Nucleic acids research}, 32:D267--D270.

\bibitem[{Bollacker et~al.(2008)Bollacker, Evans, Paritosh, Sturge, and
  Taylor}]{freebase}
Kurt Bollacker, Colin Evans, Praveen Paritosh, Tim Sturge, and Jamie Taylor.
  2008.
\newblock \href {https://doi.org/10.1145/1376616.1376746} {Freebase: A
  collaboratively created graph database for structuring human knowledge}.
\newblock In \emph{Proceedings of the 2008 ACM SIGMOD International Conference
  on Management of Data}, SIGMOD '08, page 1247–1250, New York, NY, USA.
  Association for Computing Machinery.

\bibitem[{Bordes et~al.(2014{\natexlab{a}})Bordes, Chopra, and
  Weston}]{kg_in_qa1}
Antoine Bordes, Sumit Chopra, and Jason Weston. 2014{\natexlab{a}}.
\newblock \href {https://doi.org/10.3115/v1/D14-1067} {Question answering with
  subgraph embeddings}.
\newblock In \emph{Proceedings of the 2014 Conference on Empirical Methods in
  Natural Language Processing (EMNLP)}, pages 615--620. Association for
  Computational Linguistics.

\bibitem[{Bordes et~al.(2013)Bordes, Usunier, Garcia-Duran, Weston, and
  Yakhnenko}]{bordes2013translating}
Antoine Bordes, Nicolas Usunier, Alberto Garcia-Duran, Jason Weston, and Oksana
  Yakhnenko. 2013.
\newblock Translating embeddings for modeling multi-relational data.
\newblock In \emph{Advances in neural information processing systems}, pages
  2787--2795.

\bibitem[{Bordes et~al.(2014{\natexlab{b}})Bordes, Weston, and
  Usunier}]{kg_in_qa2}
Antoine Bordes, Jason Weston, and Nicolas Usunier. 2014{\natexlab{b}}.
\newblock Open question answering with weakly supervised embedding models.
\newblock In \emph{Machine Learning and Knowledge Discovery in Databases},
  pages 165--180, Berlin, Heidelberg. Springer Berlin Heidelberg.

\bibitem[{Dettmers et~al.(2018)Dettmers, Pasquale, Pontus, and
  Riedel}]{dettmers2018conve}
Tim Dettmers, Minervini Pasquale, Stenetorp Pontus, and Sebastian Riedel. 2018.
\newblock \href {https://arxiv.org/abs/1707.01476} {Convolutional 2d knowledge
  graph embeddings}.
\newblock In \emph{Proceedings of the 32th AAAI Conference on Artificial
  Intelligence}, pages 1811--1818.

\bibitem[{Devlin et~al.(2019)Devlin, Chang, Lee, and
  Toutanova}]{devlin-etal-2019-bert}
Jacob Devlin, Ming-Wei Chang, Kenton Lee, and Kristina Toutanova. 2019.
\newblock \href {https://doi.org/10.18653/v1/N19-1423} {{BERT}: Pre-training of
  deep bidirectional transformers for language understanding}.
\newblock In \emph{Proceedings of the 2019 Conference of the North {A}merican
  Chapter of the Association for Computational Linguistics: Human Language
  Technologies, Volume 1 (Long and Short Papers)}, pages 4171--4186,
  Minneapolis, Minnesota. Association for Computational Linguistics.

\bibitem[{Dong et~al.(2014)Dong, Gabrilovich, Heitz, Horn, Lao, Murphy,
  Strohmann, Sun, and Zhang}]{kg_incomplete}
Xin Dong, Evgeniy Gabrilovich, Geremy Heitz, Wilko Horn, Ni~Lao, Kevin Murphy,
  Thomas Strohmann, Shaohua Sun, and Wei Zhang. 2014.
\newblock \href {https://doi.org/10.1145/2623330.2623623} {Knowledge vault: A
  web-scale approach to probabilistic knowledge fusion}.
\newblock In \emph{Proceedings of the 20th ACM SIGKDD International Conference
  on Knowledge Discovery and Data Mining}, KDD '14, pages 601--610, New York,
  NY, USA. ACM.

\bibitem[{Donnelly(2006)}]{snomed}
Kevin Donnelly. 2006.
\newblock {SNOMED-CT}: The advanced terminology and coding system for
  {eHealth}.
\newblock \emph{Studies in health technology and informatics}, 121:279.

\bibitem[{Gu et~al.(2020)Gu, Tinn, Cheng, Lucas, Usuyama, Liu, Naumann, Gao,
  and Poon}]{pubmedbert}
Yu~Gu, Robert Tinn, Hao Cheng, Michael Lucas, Naoto Usuyama, Xiaodong Liu,
  Tristan Naumann, Jianfeng Gao, and Hoifung Poon. 2020.
\newblock Domain-specific language model pretraining for biomedical natural
  language processing.
\newblock \emph{ArXiv}, abs/2007.15779.

\bibitem[{Guo et~al.(2020)Guo, Fan, Pang, Yang, Ai, Zamani, Wu, Croft, and
  Cheng}]{ranking}
Jiafeng Guo, Yixing Fan, Liang Pang, Liu Yang, Qingyao Ai, Hamed Zamani, Chen
  Wu, W.~Bruce Croft, and Xueqi Cheng. 2020.
\newblock \href {https://doi.org/https://doi.org/10.1016/j.ipm.2019.102067} {A
  deep look into neural ranking models for information retrieval}.
\newblock \emph{Information Processing \& Management}, 57(6):102067.

\bibitem[{{He} et~al.(2015){He}, {Zhang}, {Ren}, and {Sun}}]{prelu}
K.~{He}, X.~{Zhang}, S.~{Ren}, and J.~{Sun}. 2015.
\newblock \href {https://doi.org/10.1109/ICCV.2015.123} {Delving deep into
  rectifiers: Surpassing human-level performance on imagenet classification}.
\newblock In \emph{2015 IEEE International Conference on Computer Vision
  (ICCV)}, pages 1026--1034.

\bibitem[{{He} et~al.(2016){He}, {Zhang}, {Ren}, and {Sun}}]{resnet}
K.~{He}, X.~{Zhang}, S.~{Ren}, and J.~{Sun}. 2016.
\newblock \href {https://doi.org/10.1109/CVPR.2016.90} {Deep residual learning
  for image recognition}.
\newblock In \emph{2016 IEEE Conference on Computer Vision and Pattern
  Recognition (CVPR)}, pages 770--778.

\bibitem[{He et~al.(2016)He, Zhang, Ren, and Sun}]{preactivation}
Kaiming He, Xiangyu Zhang, Shaoqing Ren, and Jian Sun. 2016.
\newblock Identity mappings in deep residual networks.
\newblock \emph{2016 European Conference on Computer Vision (ECCV)}.

\bibitem[{Hinton et~al.(2015)Hinton, Vinyals, and Dean}]{hinton-kd}
Geoffrey Hinton, Oriol Vinyals, and Jeffrey Dean. 2015.
\newblock \href {http://arxiv.org/abs/1503.02531} {Distilling the knowledge in
  a neural network}.
\newblock In \emph{NIPS Deep Learning and Representation Learning Workshop}.

\bibitem[{Holm(1979)}]{holm}
Sture Holm. 1979.
\newblock \href {http://www.jstor.org/stable/4615733} {A simple sequentially
  rejective multiple test procedure}.
\newblock \emph{Scandinavian Journal of Statistics}, 6(2):65--70.

\bibitem[{Huang et~al.(2017)Huang, Liu, van~der Maaten, and
  Weinberger}]{huang2017densely}
Gao Huang, Zhuang Liu, Laurens van~der Maaten, and Kilian~Q Weinberger. 2017.
\newblock Densely connected convolutional networks.
\newblock In \emph{Proceedings of the IEEE Conference on Computer Vision and
  Pattern Recognition}.

\bibitem[{Huang et~al.(2019)Huang, Liu, Pleiss, Van Der~Maaten, and
  Weinberger}]{huang2019convolutional}
Gao Huang, Zhuang Liu, Geoff Pleiss, Laurens Van Der~Maaten, and Kilian
  Weinberger. 2019.
\newblock Convolutional networks with dense connectivity.
\newblock \emph{IEEE Transactions on Pattern Analysis and Machine
  Intelligence}.

\bibitem[{Jiang and Chute(2009)}]{10.1197/jamia.M2541}
Guoqian Jiang and Christopher~G. Chute. 2009.
\newblock \href {https://doi.org/10.1197/jamia.M2541} {{Auditing the Semantic
  Completeness of SNOMED CT Using Formal Concept Analysis}}.
\newblock \emph{Journal of the American Medical Informatics Association},
  16(1):89--102.

\bibitem[{Jiang et~al.(2020)Jiang, Xu, Araki, and
  Neubig}]{jiang-etal-2020-know}
Zhengbao Jiang, Frank~F. Xu, Jun Araki, and Graham Neubig. 2020.
\newblock \href {https://doi.org/10.1162/tacl_a_00324} {How can we know what
  language models know?}
\newblock \emph{Transactions of the Association for Computational Linguistics},
  8:423--438.

\bibitem[{Jim{\'e}nez-Ruiz et~al.(2011)Jim{\'e}nez-Ruiz, Grau, Horrocks, and
  Berlanga}]{Jimenez-Ruiz2011}
Ernesto Jim{\'e}nez-Ruiz, Bernardo~Cuenca Grau, Ian Horrocks, and Rafael
  Berlanga. 2011.
\newblock \href {https://doi.org/10.1186/2041-1480-2-S1-S2} {Logic-based
  assessment of the compatibility of {UMLS} ontology sources}.
\newblock \emph{Journal of Biomedical Semantics}, 2(1):S2.

\bibitem[{Kingma and Ba(2015)}]{adam}
Diederik~P. Kingma and Jimmy Ba. 2015.
\newblock \href {http://arxiv.org/abs/1412.6980} {Adam: {A} method for
  stochastic optimization}.
\newblock In \emph{3rd International Conference on Learning Representations,
  {ICLR} 2015, San Diego, CA, USA, May 7-9, 2015, Conference Track
  Proceedings}.

\bibitem[{Krizhevsky et~al.(2012)Krizhevsky, Sutskever, and Hinton}]{deepconv}
Alex Krizhevsky, Ilya Sutskever, and Geoffrey~E. Hinton. 2012.
\newblock Imagenet classification with deep convolutional neural networks.
\newblock In \emph{Proceedings of the 25th International Conference on Neural
  Information Processing Systems - Volume 1}, NIPS'12, page 1097–1105, Red
  Hook, NY, USA. Curran Associates Inc.

\bibitem[{Li et~al.(2016)Li, Taheri, Tu, and Gimpel}]{li-etal-2016-commonsense}
Xiang Li, Aynaz Taheri, Lifu Tu, and Kevin Gimpel. 2016.
\newblock \href {https://doi.org/10.18653/v1/P16-1137} {Commonsense knowledge
  base completion}.
\newblock In \emph{Proceedings of the 54th Annual Meeting of the Association
  for Computational Linguistics (Volume 1: Long Papers)}, pages 1445--1455,
  Berlin, Germany. Association for Computational Linguistics.

\bibitem[{{Li} et~al.(2017){Li}, {Yang}, {Song}, {Cao}, {Luo}, and
  {Li}}]{noisy-kd}
Y.~{Li}, J.~{Yang}, Y.~{Song}, L.~{Cao}, J.~{Luo}, and L.~{Li}. 2017.
\newblock \href {https://doi.org/10.1109/ICCV.2017.211} {Learning from noisy
  labels with distillation}.
\newblock In \emph{2017 IEEE International Conference on Computer Vision
  (ICCV)}, pages 1928--1936.

\bibitem[{Liu et~al.(2017)Liu, Wu, and Yang}]{analogy}
Hanxiao Liu, Yuexin Wu, and Yiming Yang. 2017.
\newblock \href {http://proceedings.mlr.press/v70/liu17d.html} {Analogical
  inference for multi-relational embeddings}.
\newblock In \emph{Proceedings of the 34th International Conference on Machine
  Learning}, volume~70 of \emph{Proceedings of Machine Learning Research},
  pages 2168--2178, International Convention Centre, Sydney, Australia. PMLR.

\bibitem[{Loshchilov and Hutter(2019)}]{adamw}
Ilya Loshchilov and Frank Hutter. 2019.
\newblock \href {https://openreview.net/forum?id=Bkg6RiCqY7} {Decoupled weight
  decay regularization}.
\newblock In \emph{International Conference on Learning Representations}.

\bibitem[{{Ma} et~al.(2015){Ma}, {Crook}, {Sarikaya}, and
  {Fosler-Lussier}}]{kg_in_dialog}
Y.~{Ma}, P.~A. {Crook}, R.~{Sarikaya}, and E.~{Fosler-Lussier}. 2015.
\newblock \href {https://doi.org/10.1109/ICASSP.2015.7178992} {Knowledge graph
  inference for spoken dialog systems}.
\newblock In \emph{2015 IEEE International Conference on Acoustics, Speech and
  Signal Processing (ICASSP)}, pages 5346--5350.

\bibitem[{Malaviya et~al.(2020)Malaviya, Bhagavatula, Bosselut, and
  Choi}]{malaviya2020commonsense}
Chaitanya Malaviya, Chandra Bhagavatula, Antoine Bosselut, and Yejin Choi.
  2020.
\newblock Commonsense knowledge base completion with structural and semantic
  context.
\newblock \emph{Proceedings of the 34th AAAI Conference on Artificial
  Intelligence}.

\bibitem[{Matsubara et~al.(2020)Matsubara, Vu, and Moschitti}]{qa_reranking}
Yoshitomo Matsubara, Thuy Vu, and Alessandro Moschitti. 2020.
\newblock \href {https://doi.org/10.1145/3397271.3401266} {\emph{Reranking for
  Efficient Transformer-Based Answer Selection}}, page 1577–1580. Association
  for Computing Machinery, New York, NY, USA.

\bibitem[{Mintz et~al.(2009)Mintz, Bills, Snow, and Jurafsky}]{ds_begin}
Mike Mintz, Steven Bills, Rion Snow, and Daniel Jurafsky. 2009.
\newblock \href {https://www.aclweb.org/anthology/P09-1113} {Distant
  supervision for relation extraction without labeled data}.
\newblock In \emph{Proceedings of the Joint Conference of the 47th Annual
  Meeting of the {ACL} and the 4th International Joint Conference on Natural
  Language Processing of the {AFNLP}}, pages 1003--1011, Suntec, Singapore.
  Association for Computational Linguistics.

\bibitem[{Nair and Hinton(2010)}]{relu}
Vinod Nair and Geoffrey~E. Hinton. 2010.
\newblock \href {http://dl.acm.org/citation.cfm?id=3104322.3104425} {Rectified
  linear units improve restricted boltzmann machines}.
\newblock In \emph{Proceedings of the 27th International Conference on
  International Conference on Machine Learning}, ICML'10, pages 807--814, USA.
  Omnipress.

\bibitem[{Nickel et~al.(2016)Nickel, Rosasco, and Poggio}]{hole}
Maximilian Nickel, Lorenzo Rosasco, and Tomaso Poggio. 2016.
\newblock \href {http://dl.acm.org/citation.cfm?id=3016100.3016172}
  {Holographic embeddings of knowledge graphs}.
\newblock In \emph{Proceedings of the Thirtieth AAAI Conference on Artificial
  Intelligence}, AAAI'16, pages 1955--1961. AAAI Press.

\bibitem[{Nickel et~al.(2011)Nickel, Tresp, and Kriegel}]{rascal}
Maximilian Nickel, Volker Tresp, and Hans-Peter Kriegel. 2011.
\newblock A three-way model for collective learning on multi-relational data.
\newblock In \emph{Proceedings of the 28th International Conference on
  International Conference on Machine Learning}, ICML'11, page 809–816,
  Madison, WI, USA. Omnipress.

\bibitem[{Nogueira and Cho(2019)}]{nogueira2019passage}
Rodrigo Nogueira and Kyunghyun Cho. 2019.
\newblock Passage re-ranking with {BERT}.
\newblock \emph{arXiv preprint arXiv:1901.04085}.

\bibitem[{Pei et~al.(2019)Pei, Zhang, Zhang, Sun, Lin, Sun, Wu, Jiang, Ge, Ou,
  and Pei}]{personal-rerank}
Changhua Pei, Yi~Zhang, Yongfeng Zhang, Fei Sun, Xiao Lin, Hanxiao Sun, Jian
  Wu, Peng Jiang, Junfeng Ge, Wenwu Ou, and Dan Pei. 2019.
\newblock \href {https://doi.org/10.1145/3298689.3347000} {Personalized
  re-ranking for recommendation}.
\newblock In \emph{Proceedings of the 13th ACM Conference on Recommender
  Systems}, RecSys '19, page 3–11, New York, NY, USA. Association for
  Computing Machinery.

\bibitem[{Peters et~al.(2019)Peters, Ruder, and Smith}]{peters-etal-2019-tune}
Matthew~E. Peters, Sebastian Ruder, and Noah~A. Smith. 2019.
\newblock \href {https://doi.org/10.18653/v1/W19-4302} {To tune or not to tune?
  adapting pretrained representations to diverse tasks}.
\newblock In \emph{Proceedings of the 4th Workshop on Representation Learning
  for NLP (RepL4NLP-2019)}, pages 7--14, Florence, Italy. Association for
  Computational Linguistics.

\bibitem[{Petroni et~al.(2019)Petroni, Rockt{\"a}schel, Riedel, Lewis, Bakhtin,
  Wu, and Miller}]{petroni-etal-2019-language}
Fabio Petroni, Tim Rockt{\"a}schel, Sebastian Riedel, Patrick Lewis, Anton
  Bakhtin, Yuxiang Wu, and Alexander Miller. 2019.
\newblock \href {https://doi.org/10.18653/v1/D19-1250} {Language models as
  knowledge bases?}
\newblock In \emph{Proceedings of the 2019 Conference on Empirical Methods in
  Natural Language Processing and the 9th International Joint Conference on
  Natural Language Processing (EMNLP-IJCNLP)}, pages 2463--2473, Hong Kong,
  China. Association for Computational Linguistics.

\bibitem[{Pujara et~al.(2017)Pujara, Augustine, and Getoor}]{pujara_sparsity}
Jay Pujara, Eriq Augustine, and Lise Getoor. 2017.
\newblock \href {https://doi.org/10.18653/v1/D17-1184} {Sparsity and noise:
  Where knowledge graph embeddings fall short}.
\newblock In \emph{Proceedings of the 2017 Conference on Empirical Methods in
  Natural Language Processing}, pages 1751--1756, Copenhagen, Denmark.
  Association for Computational Linguistics.

\bibitem[{Rebele et~al.(2016)Rebele, Suchanek, Hoffart, Biega, Kuzey, and
  Weikum}]{rebele2016yago}
Thomas Rebele, Fabian Suchanek, Johannes Hoffart, Joanna Biega, Erdal Kuzey,
  and Gerhard Weikum. 2016.
\newblock Yago: A multilingual knowledge base from wikipedia, wordnet, and
  geonames.
\newblock In \emph{International semantic web conference}, pages 177--185.
  Springer.

\bibitem[{Rogers et~al.(2020)Rogers, Kovaleva, and Rumshisky}]{bertology}
Anna Rogers, O.~Kovaleva, and Anna Rumshisky. 2020.
\newblock A primer in bertology: What we know about how bert works.
\newblock \emph{ArXiv}, abs/2002.12327.

\bibitem[{Ruffinelli et~al.(2020)Ruffinelli, Broscheit, and
  Gemulla}]{Ruffinelli2020You}
Daniel Ruffinelli, Samuel Broscheit, and Rainer Gemulla. 2020.
\newblock \href {https://openreview.net/forum?id=BkxSmlBFvr} {You can teach an
  old dog new tricks! on training knowledge graph embeddings}.
\newblock In \emph{International Conference on Learning Representations}.

\bibitem[{Shang et~al.(2018)Shang, Tang, Huang, Bi, He, and Zhou}]{convtranse}
Chao Shang, Yun Tang, Jing Huang, Jinbo Bi, Xiaodong He, and Bowen Zhou. 2018.
\newblock \href {http://arxiv.org/abs/1811.04441} {End-to-end structure-aware
  convolutional networks for knowledge base completion}.
\newblock \emph{CoRR}, abs/1811.04441.

\bibitem[{Simonyan and Zisserman(2015)}]{vgg}
Karen Simonyan and Andrew Zisserman. 2015.
\newblock \href {http://arxiv.org/abs/1409.1556} {Very deep convolutional
  networks for large-scale image recognition}.
\newblock In \emph{3rd International Conference on Learning Representations,
  {ICLR} 2015, San Diego, CA, USA, May 7-9, 2015, Conference Track
  Proceedings}.

\bibitem[{Socher et~al.(2013)Socher, Chen, Manning, and
  Ng}]{neural_tensor_network}
Richard Socher, Danqi Chen, Christopher~D. Manning, and Andrew~Y. Ng. 2013.
\newblock \href {http://dl.acm.org/citation.cfm?id=2999611.2999715} {Reasoning
  with neural tensor networks for knowledge base completion}.
\newblock In \emph{Proceedings of the 26th International Conference on Neural
  Information Processing Systems - Volume 1}, NIPS'13, pages 926--934, USA.
  Curran Associates Inc.

\bibitem[{Speer and Havasi(2013)}]{ConceptNet}
Robyn Speer and Catherine Havasi. 2013.
\newblock \href {https://doi.org/10.1007/978-3-642-35085-6_6} {\emph{ConceptNet
  5: A Large Semantic Network for Relational Knowledge}}, pages 161--176.
  Springer Berlin Heidelberg, Berlin, Heidelberg.

\bibitem[{Sun et~al.(2019)Sun, Deng, Nie, and Tang}]{sun2018rotate}
Zhiqing Sun, Zhi-Hong Deng, Jian-Yun Nie, and Jian Tang. 2019.
\newblock \href {https://openreview.net/forum?id=HkgEQnRqYQ} {Rotate: Knowledge
  graph embedding by relational rotation in complex space}.
\newblock In \emph{International Conference on Learning Representations}.

\bibitem[{{Tompson} et~al.(2015){Tompson}, {Goroshin}, {Jain}, {LeCun}, and
  {Bregler}}]{dropout2d}
J.~{Tompson}, R.~{Goroshin}, A.~{Jain}, Y.~{LeCun}, and C.~{Bregler}. 2015.
\newblock \href {https://doi.org/10.1109/CVPR.2015.7298664} {Efficient object
  localization using convolutional networks}.
\newblock In \emph{2015 IEEE Conference on Computer Vision and Pattern
  Recognition (CVPR)}, pages 648--656.

\bibitem[{Toutanova and Chen(2015)}]{FB15K-237}
Kristina Toutanova and Danqi Chen. 2015.
\newblock \href {https://doi.org/10.18653/v1/W15-4007} {Observed versus latent
  features for knowledge base and text inference}.
\newblock In \emph{Proceedings of the 3rd Workshop on Continuous Vector Space
  Models and their Compositionality}, pages 57--66, Beijing, China. Association
  for Computational Linguistics.

\bibitem[{Trouillon et~al.(2016)Trouillon, Welbl, Riedel, Gaussier, and
  Bouchard}]{complex}
Th{\'e}o Trouillon, Johannes Welbl, Sebastian Riedel, \'{E}ric Gaussier, and
  Guillaume Bouchard. 2016.
\newblock \href {http://dl.acm.org/citation.cfm?id=3045390.3045609} {Complex
  embeddings for simple link prediction}.
\newblock In \emph{Proceedings of the 33rd International Conference on
  International Conference on Machine Learning - Volume 48}, ICML'16, pages
  2071--2080. JMLR.org.

\bibitem[{Vashishth et~al.(2018)Vashishth, Joshi, Prayaga, Bhattacharyya, and
  Talukdar}]{reside}
Shikhar Vashishth, Rishabh Joshi, Sai~Suman Prayaga, Chiranjib Bhattacharyya,
  and Partha Talukdar. 2018.
\newblock \href {http://aclweb.org/anthology/D18-1157} {Reside: Improving
  distantly-supervised neural relation extraction using side information}.
\newblock In \emph{Proceedings of the 2018 Conference on Empirical Methods in
  Natural Language Processing}, pages 1257--1266. Association for Computational
  Linguistics.

\bibitem[{Vashishth et~al.(2020{\natexlab{a}})Vashishth, Sanyal, Nitin,
  Agrawal, and Talukdar}]{interacte2020}
Shikhar Vashishth, Soumya Sanyal, Vikram Nitin, Nilesh Agrawal, and Partha
  Talukdar. 2020{\natexlab{a}}.
\newblock \href {https://aaai.org/ojs/index.php/AAAI/article/view/5694}
  {Interacte: Improving convolution-based knowledge graph embeddings by
  increasing feature interactions}.
\newblock In \emph{Proceedings of the 34th AAAI Conference on Artificial
  Intelligence}, pages 3009--3016. AAAI Press.

\bibitem[{Vashishth et~al.(2020{\natexlab{b}})Vashishth, Sanyal, Nitin, and
  Talukdar}]{compgcn}
Shikhar Vashishth, Soumya Sanyal, Vikram Nitin, and Partha Talukdar.
  2020{\natexlab{b}}.
\newblock \href {https://openreview.net/forum?id=BylA_C4tPr} {Composition-based
  multi-relational graph convolutional networks}.
\newblock In \emph{International Conference on Learning Representations}.

\bibitem[{Wang et~al.(2011)Wang, Lin, and Metzler}]{cascade_reranking}
Lidan Wang, Jimmy Lin, and Donald Metzler. 2011.
\newblock \href {https://doi.org/10.1145/2009916.2009934} {A cascade ranking
  model for efficient ranked retrieval}.
\newblock In \emph{Proceedings of the 34th International ACM SIGIR Conference
  on Research and Development in Information Retrieval}, SIGIR '11, page
  105–114, New York, NY, USA. Association for Computing Machinery.

\bibitem[{Wang et~al.(2018)Wang, Huang, Lin, Hu, Zeng, and
  Sun}]{omnidirectionalwang2018}
Tsun-Hsuan Wang, Hung-Jui Huang, Juan-Ting Lin, Chan-Wei Hu, Kuo-Hao Zeng, and
  Min Sun. 2018.
\newblock Omnidirectional {CNN} for visual place recognition and navigation.
\newblock \emph{arXiv preprint arXiv:1803.04228}.

\bibitem[{Wolf et~al.(2020)Wolf, Debut, Sanh, Chaumond, Delangue, Moi, Cistac,
  Rault, Louf, Funtowicz, Davison, Shleifer, von Platen, Ma, Jernite, Plu, Xu,
  Le~Scao, Gugger, Drame, Lhoest, and Rush}]{wolf-etal-2020-transformers}
Thomas Wolf, Lysandre Debut, Victor Sanh, Julien Chaumond, Clement Delangue,
  Anthony Moi, Pierric Cistac, Tim Rault, Remi Louf, Morgan Funtowicz, Joe
  Davison, Sam Shleifer, Patrick von Platen, Clara Ma, Yacine Jernite, Julien
  Plu, Canwen Xu, Teven Le~Scao, Sylvain Gugger, Mariama Drame, Quentin Lhoest,
  and Alexander Rush. 2020.
\newblock \href {https://doi.org/10.18653/v1/2020.emnlp-demos.6} {Transformers:
  State-of-the-art natural language processing}.
\newblock In \emph{Proceedings of the 2020 Conference on Empirical Methods in
  Natural Language Processing: System Demonstrations}, pages 38--45, Online.
  Association for Computational Linguistics.

\bibitem[{Xiao et~al.(2016)Xiao, Huang, and Zhu}]{Xiao2016SSP}
Han Xiao, Minlie Huang, and Xiaoyan Zhu. 2016.
\newblock Ssp: Semantic space projection for knowledge graph embedding with
  text descriptions.

\bibitem[{Xie et~al.(2016)Xie, Liu, Jia, Luan, and Sun}]{xie2016}
Ruobing Xie, Zhiyuan Liu, Jia Jia, Huanbo Luan, and Maosong Sun. 2016.
\newblock Representation learning of knowledge graphs with entity descriptions.
\newblock In \emph{The 30th AAAI Conference on Artificial Intelligence}.

\bibitem[{Yang et~al.(2015)Yang, Yih, He, Gao, and Deng}]{distmult}
Bishan Yang, Scott Wen-tau Yih, Xiaodong He, Jianfeng Gao, and Li~Deng. 2015.
\newblock \href
  {https://www.microsoft.com/en-us/research/publication/embedding-entities-and-relations-for-learning-and-inference-in-knowledge-bases/}
  {Embedding entities and relations for learning and inference in knowledge
  bases}.
\newblock In \emph{Proceedings of the International Conference on Learning
  Representations (ICLR) 2015}.

\bibitem[{Yao et~al.(2019)Yao, Mao, and Luo}]{kg-bert}
Liang Yao, Chengsheng Mao, and Yuan Luo. 2019.
\newblock \href {http://arxiv.org/abs/1909.03193} {{KG-BERT:} {BERT} for
  knowledge graph completion}.
\newblock \emph{CoRR}, abs/1909.03193.

\bibitem[{Zhang et~al.(2016)Zhang, Yuan, Lian, Xie, and Ma}]{kb_recommender}
Fuzheng Zhang, Nicholas~Jing Yuan, Defu Lian, Xing Xie, and Wei-Ying Ma. 2016.
\newblock \href {https://doi.org/10.1145/2939672.2939673} {Collaborative
  knowledge base embedding for recommender systems}.
\newblock In \emph{Proceedings of the 22Nd ACM SIGKDD International Conference
  on Knowledge Discovery and Data Mining}, KDD '16, pages 353--362, New York,
  NY, USA. ACM.

\end{thebibliography}

\appendix

\section{Implementation Details}
\label{sec:implementation_details}
\subsection{BERT MLM Pre-training}
We utilize the HuggingFace Transformers library \citep{wolf-etal-2020-transformers} to work with pre-trained language models. We fine-tune the pre-trained language model with the masked-language-modeling objective upon the set of textual entity identifiers for the knowledge graph. We train the model for 3 epochs with a batch size of 32 using a learning rate of 3e-5. We use a warmup proportion of $0.1$ of the total training steps for each dataset. We use a max sequence length of 64 during this pre-training except when using the textual descriptions associated with \datafb{} where we use a max sequence length of 256. We utilize these dataset-specific language models for both generating the entity embeddings and for initializing the re-ranking model.

\subsection{Ranking}

\subsubsection{Training Procedure}
We train all of the ranking models implemented in this work for a maximum of $200$ epochs and terminate training early if the validation MRR has not improved for $20$ epochs. For evaluation, we reload the model weights from the epoch that achieved the best validation MRR and evaluate it on the test set.

\subsubsection{BERT-ResNet Implementations}
For our BERT-ResNet model, we set $f=5$ where $f$ is the hyperparameter that controls the size of the spatial feature map produced by the initial 1D convolution. Thus our initial 1D convolution has $f\times f=25$ filters. We set $N=2$ where $N$ is the hyperparameter that controls the depth of the convolutional network. This means that our BERT-ResNet model consists of $3N=6$ sequential bottleneck blocks. 

We trained the models using a batch size of $64$ with a 1vsAll strategy \citep{Ruffinelli2020You} with the binary cross entropy loss function. We use the Adam optimizer \citep{adam} with decoupled weight decay regularization \citep{adamw} and train the model with a learning rate of 1e-3. We use label smoothing with a value of 0.1, clip gradients to a max value of 1, and regularize the model using weight decay with a weight of 1e-4. We apply dropout with drop probability $0.2$ after the embedding layer and apply 2D dropout \citep{dropout2d} with the same drop probability before the 2D convolutions. We apply dropout with probability $0.3$ after the pooling and fully connected layer. We manually tuned the hyperparameters for this model based on validation performance.

\subsubsection{Baseline Implementations}
For our baseline implementations of DistMult, ComplEx, ConvE, and ConvTransE, we adapt the implementations released by \citet{dettmers2018conve} and \citet{malaviya2020commonsense}. We utilize the hyperparameters reported in the original papers and conduct a grid search to tune the embedding dimension from [100, 200, 300] and the initial learning rate from [5e-3, 1e-3, 5e-4, 1e-4] for each dataset. We train the models with a batch size of 128 using the 1vsAll strategy with the cross entropy loss function because the replication study by \citet{Ruffinelli2020You} found that this training strategy generally led to better performance than other training strategies. For the grid search, we train each model for a maximum of $50$ epochs and then select the hyperparameters with the best validation performance and retrain the model with our aforementioned training procedure.

For our implementation of BERT-ConvE and BERT-ConvTransE, we adapt the baseline ConvE and ConvTransE to use BERT embeddings in the same manner as our model. The convolution for BERT-ConvE has $32$ channels and the convolution for BERT-ConvTransE has $64$ channels. These values were selected to produce models with a comparable number of trainable parameters to our model. We then project the final feature vector down to the embedding dimensionality and rank candidates identically to our model.

We trained both models with a batch size of $64$ using 1vsAll strategy \citep{Ruffinelli2020You} with the binary cross entropy loss function using the Adam optimizer \citep{adam} with decoupled weight decay regularization \citep{adamw}. We train the models with a learning rate of 1e-4, use label smoothing with value 0.1, clip gradients to a max value of 1, and regularize the model using weight decay with a weight of $0.0001$. We apply dropout with drop probability $0.2$ after the embedding layer and after the convolution. We apply dropout with probability $0.3$ after the fully connected layer. 

For our baseline BERT-DeepConv model, we use the same hyperparamters as BERT-ResNet for the initial 1-D convolution and then apply a sequence of three $3 \times 3$ convolutions with circular padding. The second convolution doubles the number of channels so the dimensionality of the final feature map produced by the sequence of convolutions is $2d$. We then mean pool and project the feature map to the embedding dimensionality identically to our proposed model. We selected these hyperparameters so that this baseline has a similar number of trainable parameters to our proposed model. All other implementation details are identical to our BERT-Resnet model (e.g. use of pre-activations, application of dropout, training hyperparameters, etc.).

\subsection{Re-Ranking}
We fine-tune BERT with a learning rate of $3\mathrm{e}{-5}$ using the Adam optimizer \citep{adam} with decoupled weight decay regularization \citep{adamw}. We truncate the textual triple representation to a max length of $32$ tokens and fine-tune BERT with a batch size of 128 for a maximum of 10 epochs. Training is terminated early if the validation MRR does not improve for $3$ epochs. We set the weight decay parameter to $0.01$ and clip gradients to a max value of 1 during training. We apply dropout with probability $0.3$ to the final feature representation before the prediction and otherwise use the default parameters provided by the HuggingFace Transformers library \citep{wolf-etal-2020-transformers}. We set $\lambda=0.5$ for \datasmd{}, $\lambda=1.0$ for \datacon{}, and $\lambda=0.75$ for \datafb{} and \datafbs{}. We set the temparature as $T=1$ for all models.

\section{Evaluation Metrics}
We provide a mathematical formulation for our evaluation metrics. If we denote the set of all facts in the test set as $\mathcal{T}$, then the Mean Rank (MR) is simply computed as \[\textrm{MR} = \frac{1}{\vert \mathcal{T} \vert} \sum_{x_i \in \mathcal{T}}  \textrm{rank}(x_i)\]
The Mean Reciprocal Rank (MRR) is computed as \[\textrm{MRR} = \frac{1}{\vert \mathcal{T} \vert} \sum_{x_i \in \mathcal{T}}  \frac{1}{\textrm{rank}(x_i)}\] 
The Hits at k (H@k) is calculated as \[\textrm{H@k} = \frac{1}{\vert \mathcal{T} \vert} \sum_{x_i \in \mathcal{T}} I[\textrm{rank}(x_i) \leq k]\] where $I[P]$ is $1$ if the condition $P$ is true and is $0$ otherwise. When computing $\textrm{rank}(x_i)$, we first filter out all positive samples other than the target entity $x_i$. This is commonly referred to as the filtered setting.

\section{Supplementary Tables}
\label{sec:supplemental_tables_figs}

\begin{table}[!h]
\centering
\scriptsize
\begin{tabular}{lccccc}
\toprule
       & \multicolumn{5}{c}{\textbf{\datasmd{}}}  \\
\cmidrule(r){2-6}
       & MR & MRR   & H@1     & H@3 & H@10 \\
\midrule
DistMult   & $5039$  & $.294$  & $.226$   & $.319$  & $.427$  \\
ComplEx  & $3850$  & $.303$  & $.225$   & $.335$  & $.457$  \\
ConvE & $3618$  & $.271$  & $.191$   & $.303$  & $.429$  \\
ConvTransE & $3484$  & $.293$  & $.216$   & $.323$  & $.446$  \\
\midrule
BERT-ConvE & $386$  & $.384$  & $.278$   & $.431$  & $.593$  \\
BERT-ConvTransE & $487$  & $.374$  & $.274$   & $.417$  & $.569$  \\
\midrule
BERT-DeepConv & $250$  & $.481$  & $.376$   & $.534$  & $.687$  \\
\midrule
BERT-ResNet & $249$  & $.493$  & $.389$   & $.546$  & $.694$  \\ 
\bottomrule
\end{tabular}
\caption{Validation ranking results for \datasmd{}. }
\end{table}

\begin{table}[!h]
\centering
\scriptsize
\begin{tabular}{lccccc}
\toprule
       & \multicolumn{5}{c}{\textbf{\datacon{}}}  \\
\cmidrule(r){2-6}
       & MR & MRR   & H@1     & H@3 & H@10 \\
\midrule
BERT-ConvE & $283$  & $.370$ & $.253$ & $.423$  & $.606$  \\
BERT-ConvTransE & $323$  & $.381$  & $.267$   & $.430$  & $608$  \\
\midrule
BERT-DeepConv & $261$  & $.463$  & $.342$   & $.526$  & $.705$  \\
\midrule
BERT-ResNet & $269$  & $.463$  & $.341$   & $.53$  & $.700$  \\ 
\bottomrule
\end{tabular}
\caption{Validation ranking results for \datacon{}.}
\end{table}

\begin{table}[!h]
\centering
\scriptsize
\begin{tabular}{lccccc}
\toprule
       & \multicolumn{5}{c}{\textbf{\datafb{}}}  \\
\cmidrule(r){2-6}
       & MR & MRR   & H@1     & H@3 & H@10 \\
\midrule
BERT-ConvE & $189$  & $.308$  & $.228$ & $.334$ & $.467$  \\
BERT-ConvTransE & $208$ & $.301$ & $.224$ & $.326$ & $.449$  \\
\midrule
BERT-DeepConv & $186$ & $.332$ & $.251$ & $.360$ & $.490$  \\
\midrule
BERT-ResNet & $185$  & $.351$  & $.269$   & $.384$  & $.514$  \\ 
\bottomrule
\end{tabular}
\caption{Validation ranking results for \datafb{}.}
\end{table}

\begin{table}[!h]
\centering
\scriptsize
\begin{tabular}{lccccc}
\toprule
       & \multicolumn{5}{c}{\textbf{\datafbs{}}}  \\
\cmidrule(r){2-6}
       & MR & MRR   & H@1     & H@3 & H@10 \\
\midrule
DistMult   & $3034$  & $.136$  & $.093$   & $.146$  & $.227$  \\
ComplEx  & $3311$  & $.134$  & $.092$   & $.144$  & $.220$  \\
ConvE & $2247$  & $.158$  & $.107$ & $.166$  & $.261$  \\
ConvTransE & $2275$ & $.154$ & $.103$ & $.163$  & $.257$  \\
\midrule
BERT-ConvE & $412$  & $.192$  & $.128$   & $.202$  & $.321$  \\
BERT-ConvTransE & $390$  & $.192$  & $.129$   & $.204$  & $.318$  \\
\midrule
BERT-DeepConv & $419$  & $.193$  & $.131$   & $.203$  & $.320$  \\
\midrule
BERT-ResNet & $412$  & $.194$  & $.131$   & $.204$  & $.321$  \\ 
\bottomrule
\end{tabular}
\caption{Validation ranking results for \datafbs{}.}
\end{table}

\begin{table}[!h]
\centering
\scriptsize
\begin{tabular}{lcccc}
\toprule
       & \multicolumn{4}{c}{\textbf{\datasmd{}}}  \\
\cmidrule(r){2-5}
       & MR & MRR   & H@1     & H@3  \\
\midrule
BERT-ResNet   & $2$  & $.698$  & $.561$   & $.787$    \\
\hspace{3mm} + Re-ranking + KD + TE   & $2$  & $.822$  & $.724$   & $.901$    \\
\midrule
       & \multicolumn{4}{c}{\textbf{\datacon{}}}  \\
\cmidrule(r){2-5}
       & MR & MRR   & H@1     & H@3  \\
\midrule
BERT-ResNet   & $3$  & $.648$  & $.488$   & $.758$    \\
\hspace{3mm} + Re-ranking + KD + TE   & $2$  & $.668$  & $.511$   & $.780$    \\
\midrule
       & \multicolumn{4}{c}{\textbf{\datafb{}}}  \\
\cmidrule(r){2-5}
       & MR & MRR   & H@1     & H@3  \\
\midrule
BERT-ResNet   & $3$  & $.664$  & $.523$ & $.748$    \\
\hspace{3mm} + Re-ranking + KD + TE   & $3$  & $.678$  & $.539$   & $.761$    \\
\midrule
       & \multicolumn{4}{c}{\textbf{\datafbs{}}}  \\
\cmidrule(r){2-5}
       & MR & MRR   & H@1     & H@3  \\
\midrule
BERT-ResNet   & $3$  & $.567$  & $.407$   & $.634$    \\
\hspace{3mm} + Re-ranking + KD + TE   & $3$  & $.589$  & $.427$   & $.667$    \\
\bottomrule
\end{tabular}
\caption{Validation re-ranking results. We report metrics for the subset of queries where the retrieved entity is already in the top $10$ entities because the re-ranking procedure leaves other rankings unchanged. }
\end{table}

\end{document}